# Evaluating Large Language Models for Automated Clinical Abstraction in Pulmonary Embolism Registries: Performance Across Model Sizes, Versions, and Parameters

Mahmoud Alwakeel[1(✉)], Emory Buck[1], Jonathan G. Martin[1], Imran Aslam[1], Sudarshan Rajagopal[1], Jian Pei[2], Mihai V. Podgoreanu[1], Christopher J. Lindsell[3], An-Kwok Ian Wong[1]

[1] Duke University Health System, Durham, North Carolina, USA; [2] Duke University, Durham, North Carolina, USA; [3] Duke Clinical Research Institute, Durham, North Carolina, USA
{mia21,esb42,jm332,mia20,rajag007,jp579,podgo001,cjl83,akw19}@duke.edu

**Abstract.** Pulmonary embolism (PE) registries accelerate practice-improving research but depend on resource-intensive manual abstraction of radiology reports. We evaluated whether openly available large-language models (LLMs) can automate concept extraction from computed-tomography PE (CTPE) reports without sacrificing data quality. Four Llama-3 (L3) variants (3.0 8 B, 3.1 8 B, 3.1 70 B, 3.3 70 B) and two reviewer models Phi-4 (P4) 14 B and Gemma-3 27 B (G3) were tested on 250 dual-annotated CTPE reports each from MIMIC-IV and Duke University. Outcomes were accuracy, positive predictive value (PPV), and negative predictive value (NPV) versus a human gold standard across model sizes, temperature settings, and shot counts. Mean accuracy across all concepts increased with scale: 0.83 (L3-0 8 B), 0.91 (L3-1 8 B), and 0.96 for both 70 B variants; P4 14 B achieved 0.98; G3 matched. Accuracy differed by $<$ 0.03 between datasets, underscoring external robustness. In dual-model concordance analysis (L3 70 B + P4 14 B), PE-presence PPV was $\geq$ 0.95 and NPV $\geq$ 0.98, while location, thrombus burden, right-heart strain, and image-quality artifacts each maintained PPV $\geq$ 0.90 and NPV $\geq$ 0.95. Fewer than 4 % of individual concept annotations were discordant, and complete agreement was observed in more than 75 % of reports. G3 performed comparably. LLMs therefore offer a scalable, accurate solution for PE registry abstraction, and a dual-model review workflow can further safeguard data quality with minimal human oversight.

## 1 Introduction

Pulmonary embolism (PE) is a leading cause of cardiovascular mortality worldwide, and demands prompt, accurate risk stratification for optimal care.[1] The PE Response Teams (PERT) are multidisciplinary groups, cardiologists, radiologists, vascular surgeons, and intensivists that coordinate timely and effective interventions for PE patients. These interventions have been associated with reduced mortality and shorter hospital stays.[2] Recognizing the complexity and variability in PE management, the non-profit PERT Consortium has advocated for a standardized registry to document interventions, outcomes, and quality metrics.[3] Yet its reliance on manual abstraction limits scalability, completeness, and generalizability.[3]

Traditional automation—rule-based keyword searches, structured templates, and classical machine-learning models—partially eases workload but struggles to interpret nuanced language, accommodate provider variability, and integrate context in free-text documentation.[4–6] Furthermore, rule-based systems lack adaptability across institutions and electronic health records, requiring extensive site-specific customization.[7] Large language models (LLMs) use advanced natural language processing to extract clinical concepts with minimal tuning and can reach human-level performance with superior consistency and scalability.[8] However, their use in high-stakes registries such as PE remains underexplored.

This study tests publicly available LLMs on computed-tomography PE (CTPE) reports to automate PERT-registry abstraction. We evaluate multiple LLM configurations and introduce a dual-model workflow in which an independent reviewer LLM audits each extraction, providing scalable quality assurance.

## 2 Methods

### 2.1 Study Design

This retrospective study used the MIMIC-IV (2008-2019) and Duke University Health System (DUHS) databases (2014-2024). All CTPE reports for adults ≥ 18 years were retrieved. Radiologic variables followed the latest PERT-Consortium definitions.[9] Key concepts were PE detection and location (right/left; saddle, main, lobar, segmental, subsegmental arteries); right-heart strain (dilated right ventricle (RV), septal flattening, IV-contrast reflux); high thrombus burden and image quality (motion, bolus artifacts). Each variable was labeled present (1) or absent / missing / uncertain (0), mirroring PERT Consortium registry-documentation conventions, simplifying downstream analysis, and aligning with routine clinical interpretation. Two clinicians with expertise in intensive care, radiology, and cardiology independently annotated the data after a standardized training; unresolved disagreements were adjudicated by a third clinician. Figure 1 shows the workflow. We compared Llama-3 (L3) variants—3.0 8 B, 3.1 8 B, 3.1 70 B, and 3.3 70 B—using zero- to five-shot examples and temperatures 0.0 – 1.5. Generation failures (e.g., missing JSON) were tagged -99. The DUHS IRB deemed the study exempt (Pro00115368).

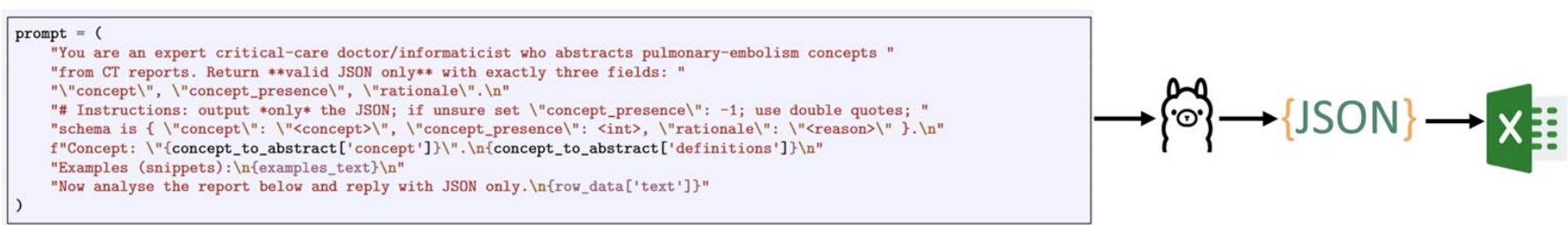

**Fig 1.** Workflow of LLM-Based Concept Extraction

### 2.2 Outcome and Statistical Analysis

The primary outcome was LLM accuracy versus a human gold standard across model sizes, versions, temperatures, and shot counts. To build quality assurance framework, Phi-4 (P4) 14 B and Gemma-3 27 B (G3) served as a reviewer LLMs. Pairwise concordance was defined as cases in which L3 and either P4 or G3 produced the same label for a given concept. These concordant calls were then compared with the human reference: joint positives yielded the positive predictive value (PPV) of the dual-model system, and joint negatives yielded the negative predictive value (NPV). Disagreement was quantified at two granularities: a record-level disagreement was logged whenever L3 and P4/G3 differed on any concept within a CTPE report, prompting a full-record review, whereas a concept-level disagreement captured only those individual concept labels on which L3 and P4/G3 diverged, limiting review to those concepts. P4/G3 ran at temperature 0, zero-shot for all tests. Continuous variables are reported as mean ± SD; categorical variables as counts (%). Sample-size planning (binomial precision method, anticipated accuracy 0.90, ±5 percentage-point two-sided 95 % CI, α 0.05) required N = 139; accordingly, we retained 250 CTPE reports per dataset to accommodate annotation drift.[10] All analyses used Python 3.11.5, following STARD guidelines. We complied with the PhysioNet LLM responsible use guidelines by not using online LLM services.[11]

## 3 Results

A total of 9,132 studies from MIMIC-IV and 82,004 from DUHS EHR were included. Of these, 250 studies from each dataset underwent dual physician abstraction to serve as the criterion standard, randomly sampled for MIMIC-IV and year-stratified (5 per year) for DUHS. On average, the 8 B and 14 B models processed a report in ≤ 30 seconds, the 70 B models in ≈ 2 minutes, whereas human reviewers required ≈ 1.5 minutes. Figure 2 presents the gold-standard concepts distribution.

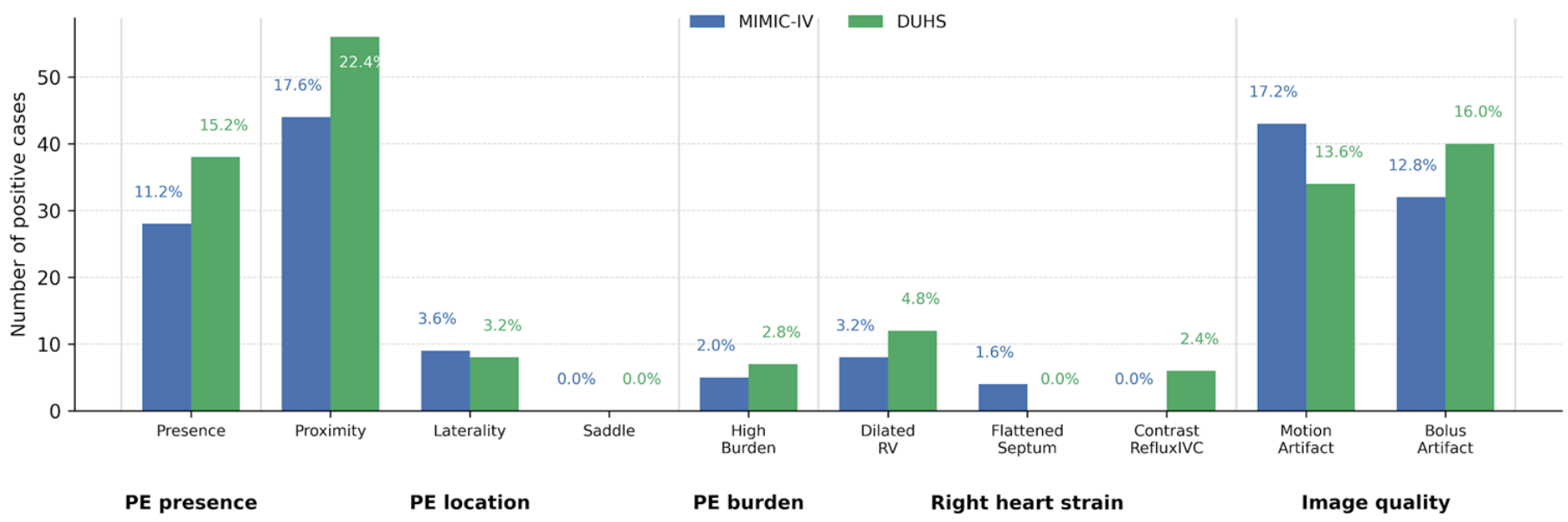

**Fig. 2.** Gold-standard positive counts by concept in MIMIC-IV vs DUHS

Mean accuracy rose with model size—0.83 for L3-0 8 B, 0.91 for L3-1 8 B, 0.96 for both 70 B variants, 0.94 for G3 27B and 0.98 for P4 14 B—with a dataset gap < 0.03 between MIMIC-IV and DUHS. The largest gains appeared in relatively infrequent right-heart-strain and thrombus-burden markers: IVC reflux climbed from 0.52 to 0.98, septal flattening from 0.73 to 1.00, high thrombus burden from 0.56 to 0.99, and saddle PE from 0.65 to 1.00. By contrast, common attributes such as laterality and proximal-vessel category were already ≥ 0.90 with L3-0 8 B and improved by no more than 0.05 with larger models. Figure 3 summarizes detailed accuracies for all CTPE concepts.

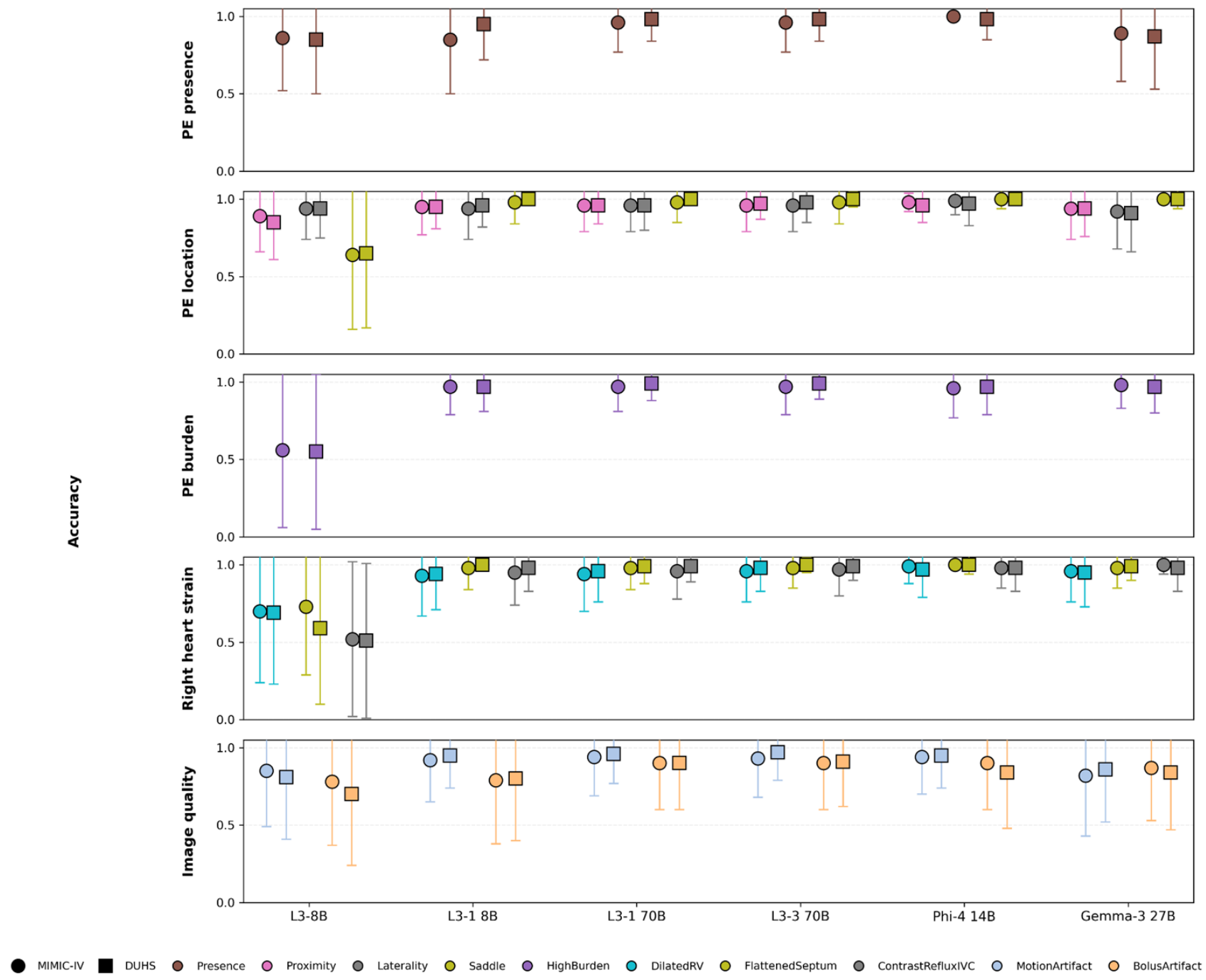

**Fig. 3**. Mean accuracy (± SD) of CTPE concepts extractions across MIMIC-IV and DUHS cohorts, stratified by sub-category and model.

Figure 4 shows model accuracy as temperature (0 – 1.5) and in-context-example count (0 – 5 shots) vary. Across both MIMIC-IV and DUHS, the L3 70 B variants remained stable at ≥0.95 accuracy under all prompt settings, demonstrating prompt-agnostic robustness. In contrast, the 8 B models were markedly prompt-sensitive: performance peaked with temperatures 0.2 – 0.5 and one to two shots, but declined by 6 – 10 percentage points—most notably for thrombus-burden and right-heart-strain concepts—when temperature exceeded 1.0 or the shot count surpassed three. The L3-1 8B partially mitigated this volatility yet continued to lag the 70 B family by roughly five percentage points. Nearly identical response curves in the two datasets indicate that these trends are attributable to model architecture rather than corpus-specific factors. Adopting a dual-model concordance strategy for quality-assurance framework, Figure 5 shows that reports on which the predictions of L3 70 B and P4 14 B/G3 27B agree achieve high performance. P4-concordant, for PE presence, both PPV and NPV approach 1.00, while for PE location, thrombus burden, right-heart strain and image-quality artifacts the concordant PPV ≥ 0.90 and the NPV remains ≥ 0.95.

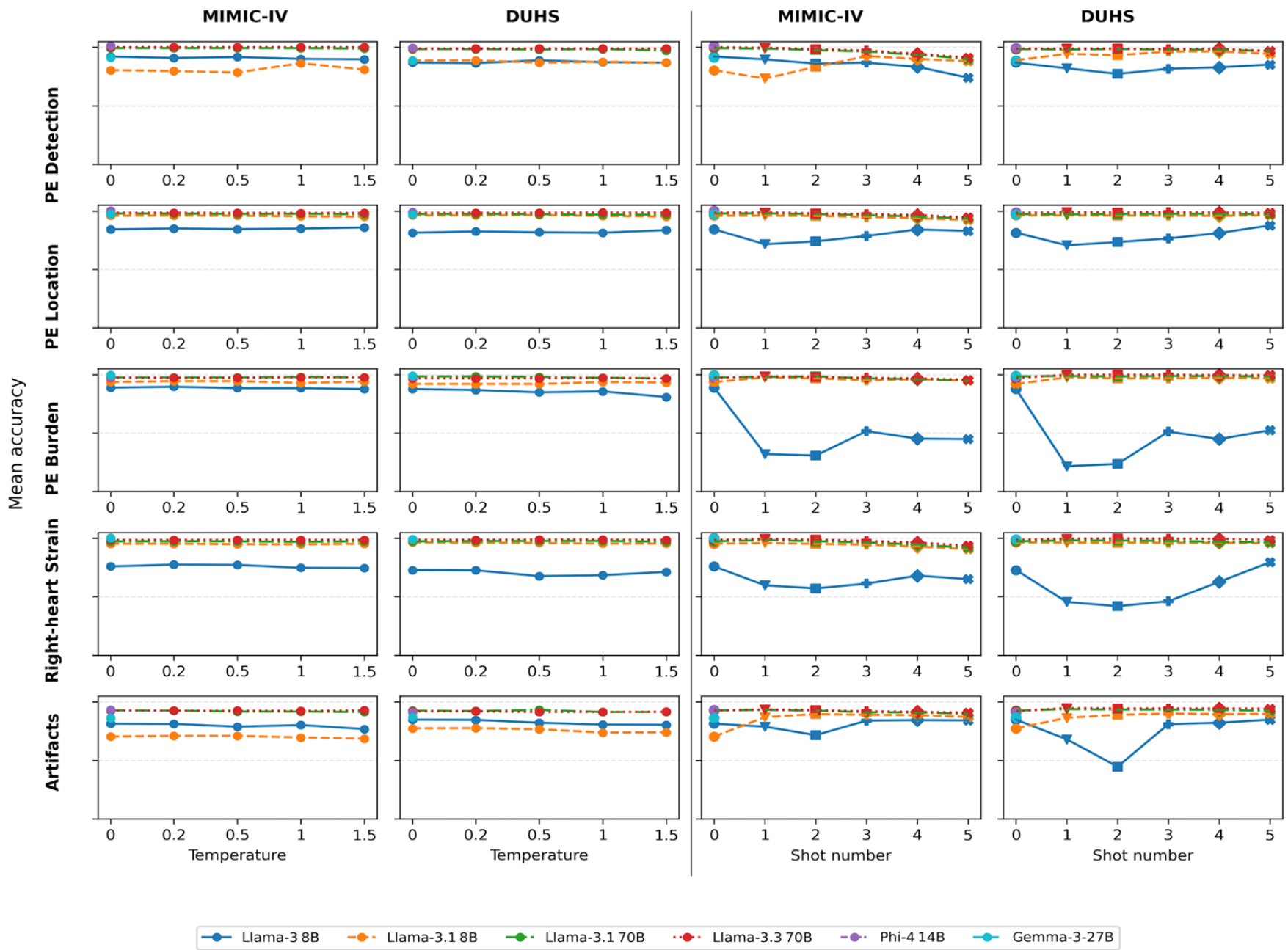

**Fig. 4**. Model accuracy across temperatures (left) and shot numbers (right) on MIMIC-IV and DUHS

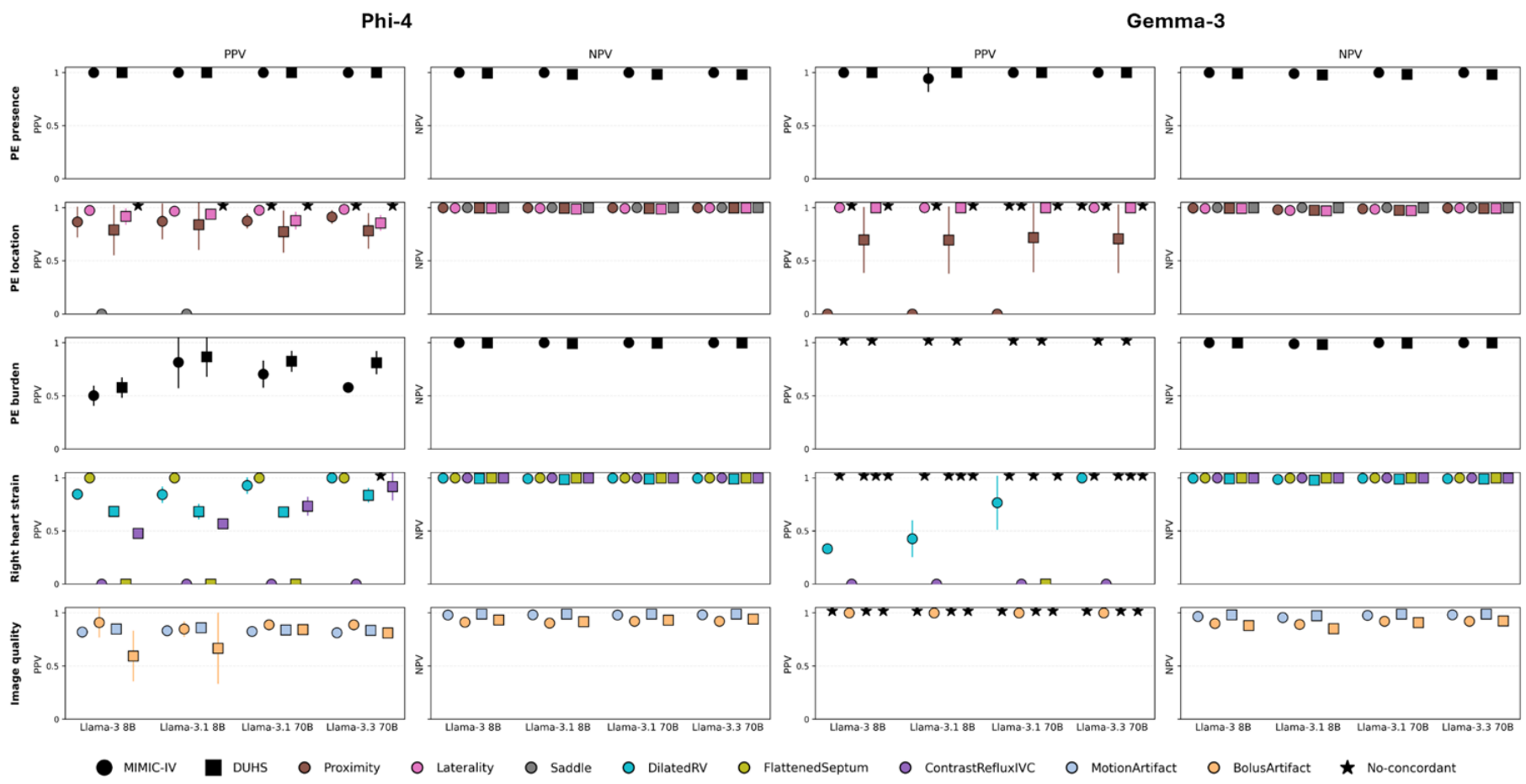

**Fig. 5.** Positive- (PPV) and negative-predictive values (NPV) for clinical concepts across models on MIMIC-IV and DUHS

Table 1 summarizes the post L3-P4/G3 concordant filter manual review workload and partitions residual cases into reviewer-correct, reviewer-misclassified, and parsing failures (from either reviewer or L3). The 70 B models cut record-level workload from 87–91% to 17–26% and concept-level workload from 20–23% to 3–4%. After filtering, parsing errors dominate the residual set for P4 whereas misclassifications dominate for G3. Consequently, once concordant outputs are accepted automatically, approximately 75 % of reports and > 96 % of individual labels require no manual review, keeping residual error below 4 %. The G3-27 B reviewer performs comparably overall but still lacks concordant positives for several low-prevalence findings (saddle embolus, high-burden PE, and certain right-heart-strain signs), leaving its record- and concept-level disagreement about five percentage points higher than the P4 pairing.

**Table 1.** L3 pipelines: P4 vs G3 reviewer outcomes and workload. Concept % = individual concepts need human review, Record % = record with at least one discordant concept; Corr = reviewer model-correct, Mis = misclassified, Parse = unparsable output

| Dataset / Model | P4 Con / Rec \| Corr / Mis / Parse | G3 Con / Rec \| Corr / Mis / Parse |
|---|---|---|
| MIMIC-IV - L3-8B | 19.7 / 87.2 \| 8.7 / 9.1 / 82.3 | 24.7 / 89.0 \| 11.8 / 22.2 / 66.0 |
| MIMIC-IV - L3-1 8B | 6.0 / 36.9 \| 49.5 / 15.2 / 35.3 | 8.9 / 48.8 \| 34.9 / 39.9 / 25.2 |
| MIMIC-IV - L3-1 70B | 3.8 / 18.6 \| 33.9 / 10.1 / 56.0 | 8.4 / 40.7 \| 27.1 / 46.2 / 26.7 |
| MIMIC-IV - L3-3 70B | 3.7 / 17.3 \| 25.7 / 16.1 / 58.2 | 8.7 / 38.8 \| 24.6 / 49.3 / 26.1 |
| DUHS - L3-8B | 23.3 / 90.5 \| 14.0 / 5.5 / 80.5 | 28.3 / 91.0 \| 14.5 / 19.0 / 66.6 |
| DUHS - L3-1 8B | 4.6 / 37.1 \| 58.1 / 39.0 / 2.9 | 6.5 / 40.0 \| 30.4 / 65.8 / 3.8 |
| DUHS - L3-1 70B | 3.4 / 26.3 \| 39.9 / 56.6 / 3.5 | 6.5 / 41.4 \| 19.3 / 77.1 / 3.6 |
| DUHS - L3-3 70B | 3.0 / 26.2 \| 23.9 / 70.3 / 5.8 | 7.4 / 42.7 \| 17.1 / 79.0 / 3.9 |

## 4 Discussion

This study shows that openly available LLMs can extract pulmonary-embolism concepts from CT angiography reports with high fidelity. The L3 70 B models achieved an overall accuracy of 0.96, out-performing the 3.1 8 B (0.91) and 3.0 8 B (0.83) variants, while the independent P4 14 B reviewer reached 0.98. Across both health-system corpora, accuracy for PE presence, location, thrombus burden, right-heart strain, and image-quality artifacts remained between 0.90 and 0.97. Dual-model concordance (L3 70 B + P4 14 B) further tightened performance: fewer than 4 % of individual concept annotations disagreed with the human reference, and full agreement was achieved in more than 75 % of reports; within this concordant subset, PPV was ≥ 0.95 and NPV ≥ 0.98 for PE presence and remained ≥ 0.90 and ≥ 0.95, respectively, for the other concept categories. Replacing the reviewer with G3 27 B preserved the ≥ 0.95 PPV/NPV envelope while increasing residual disagreement by only five percentage-points. Moderate temperature settings (0.2–0.5) with one to two in-context examples delivered the best balance between structured guidance and adaptive reasoning, and simplifying definition wording improved model–human alignment. Together, these findings indicate that large LLMs, especially when paired for concordance review, can markedly reduce manual abstraction workload while preserving registry-grade data quality.

LLMs in medical concept extraction yield promising outcomes, as evidenced by a few available studies.[12–14] Our study expands this work by evaluating multiple openly available LLMs of different parameter sizes and versions on a more complex and nuanced set of clinical concepts. Unlike prior studies that focused on well-defined categories, we tested LLMs on concepts that posed significant challenges even for human experts, requiring multiple multidisciplinary meetings and continuous feedback checks to improve agreement. By incorporating sophisticated clinical reasoning tasks, along with a dual-model concordance check, our study provides a more comprehensive assessment of LLM performance in medical abstraction. Finally, we outline a pragmatic, cost-effective quality-assurance framework for selecting LLM settings that yield high predictive fidelity, enabling automated PERT-registry abstraction while simultaneously monitoring performance thresholds so that residual errors remain low and any records flagged for uncertainty are efficiently routed for human review.

While techniques such as multi-shot prompting often improve LLM performance in general AI/ML tasks, their role in medical concept extraction remains uncertain, and prior studies have reported mixed results.[15, 16] Our findings align with this literature: multi-shot examples did not reliably increase accuracy or predictive values and occasionally degraded performance. This effect may reflect greater prompt complexity and the inherent ambiguity of clinical language, which can overload the model's reasoning capacity.[17, 18] Longer, example-heavy prompts are also known to dilute the model's attention, the "lost-in-the-middle" phenomenon, leading to similar performance drops in open-domain benchmarks.[19] Consistent with previous work, lower temperature settings (0.2–0.5) produced the best accuracy, whereas higher temperatures (> 1.0) diminished performance—particularly for nuanced concepts such as PE location and right-heart strain, where even human agreement is challenged. Finally, our two-model concordance framework markedly increased confidence in abstraction quality when both models agreed, yielding high performance PPV/NPV (≈ 0.95 / 0.98), in line with earlier reports on dual-LLM validation.[20]

This study encountered several challenges that can help guide future work on utilizing LLMs for automating databases and registries. One of the primary challenges was developing precise definitions for medical concepts, taking into account both direct and indirect findings. For example, if a report does not explicitly mention a movement artifact, how should it be annotated? A valuable approach was to hold multidisciplinary meetings involving radiologists who write the reports and clinicians involved in patient care to determine whether it was directly mentioned, not mentioned, or implied to be negative based on radiological and clinical perspectives. It is important to acknowledge that decisions about managing ambiguous or uncertain cases are generally made by the abstraction team. Maintaining records at a more granular level may be beneficial for quality monitoring, audits, or future manual reviews. In addition to our internal consensus, we integrated the latest national consortium guidelines relevant to our work to ensure future interoperability.

Our study's strength lies in its multicenter, multi-institutional design, which included both academic teaching hospitals (MIMIC and DUHS data) and community hospitals (through DUHS), thereby enhancing the generalizability and reliability of our findings. We evaluated a spectrum of openly available LLMs of different sizes and minor versions, systematically testing multi-shot prompting and temperature settings to create a comprehensive framework for performance optimization. Introducing P4 14 B as an independent reviewer enabled a dual-model concordance workflow: agreement between L3 70 B and P4 14 B was strongest for well-defined concepts like PE presence, where concordant cases showed PPV ≥ 0.95 and NPV ≥ 0.98 and covered more than 75 % of reports, leaving < 4 % residual concept-level disagreement. Variability was greatest for ambiguous findings such as right-heart strain, and record-level disagreement was modestly higher in DUHS than in MIMIC-IV; nevertheless, the 70 B models consistently maintained accuracies above 0.90, and their pairing with P4 preserved high predictive values, sustaining strong confidence in the automated abstraction pipeline. Together, these design choices lay the foundation for a scalable, fully automated monitoring system that flags uncertain or discordant records for targeted human review.

This study was limited by not testing the performance of larger LLMs, such as Llama 3.1 405 B, due to computational resource constraints within PHI-capable, institution-approved computers. Moving from 8 B to 70 B models led to an average four-fold increase in processing time, and running two large models in parallel further amplified the compute burden; however, the very high accuracy and predictive values achieved by the 70 B models suggested diminishing returns from evaluating even larger architectures. Additionally, our study focused on a predefined subset of radiologic concepts related to PE, and did not address all possible findings present in CTPE reports. This limited concept coverage may restrict the generalizability of our results to broader clinical abstraction tasks or other registry applications. Further research is needed to evaluate the models' performance across a broader range of clinical domains. Another limitation is the lack of real-time clinical validation, as our results are based on retrospective data. Furthermore, while we examined the effect of multi-shot prompting and temperature tuning, we did not fully explore alternative fine-tuning strategies. Lastly, although our study focused specifically on clinical concepts extracted from CTPE reports, the PERT Consortium registry also includes additional data elements such as echocardiogram findings, which we did not evaluate.

## 5 Conclusion

Our study shows that openly available LLMs can automate CTPE concept extraction with ≥ 0.96 accuracy and concordant PPV ≥ 0.95 / NPV ≥ 0.98, reducing manual workload to < 4 % of individual concepts when a dual-model concordance screen is applied. These results indicate that LLM-assisted abstraction can markedly improve the scale, speed, and cost-efficiency of clinical registry creation while preserving data fidelity. Future work should focus on prospective, real-time validation; extension to additional clinical domains; and adaptive quality-assurance frameworks that continuously monitor LLM performance and trigger targeted human review when thresholds are breached.

**Prospect of application:** Deployed within radiology information systems, the high-trust / low-burden Llama–Phi concordance engine auto-abstracts CTPE reports in seconds, streams near-real-time data to PERT registries / analytics dashboards, and routes only discordant records (< 25 %) for review—reducing manual effort by > 75%, compressing reporting cycles, and supporting continuous, data-driven optimization of pulmonary-embolism care.

## 6 Acknowledgment

This work was partially supported by Mark III Systems with GPU resources for de-identified data. AIW is supported by the National Heart, Lung, and Blood Institute under R01HL177003.